\documentclass[letterpaper, 10 pt, conference]{ieeeconf}  % Comment this line out if you need a4paper

\usepackage{amsmath}
\usepackage{amssymb}
\usepackage{graphicx}
\usepackage{booktabs}
\usepackage {bm}
\usepackage{xcolor}
\usepackage{amssymb}
\usepackage{caption}  % 放在导言区
\usepackage{pifont}
\usepackage{wrapfig}
\usepackage{subcaption}
\usepackage{placeins}
\usepackage[T1]{fontenc}
\usepackage{cleveref}
\usepackage{array}
\usepackage{tabularx}
\usepackage{multirow}
\usepackage{tikz}
\usepackage{url}
\usepackage{bbm}
\usepackage{fontawesome}

\IEEEoverridecommandlockouts                              % This command is only needed if 
\title{\LARGE \bf
TrajMoE: Scene-Adaptive Trajectory Planning with Mixture of Experts and Reinforcement Learning
}

\author{
Zebin Xing$^{1}$ \hspace{0.1cm} 
Pengxuan Yang$^{1}$ \hspace{0.1cm} 
Linbo Wang$^{1}$ \hspace{0.1cm} 
Yichen Zhang$^{1}$ \hspace{0.1cm}
Yiming Hu$^{1}$ \hspace{0.1cm}
Yupeng Zheng$^{\dagger 1}$ \hspace{0.1cm} \\
Junli Wang$^{1,2}$ \hspace{0.1cm}
Yinfeng Gao$^{1,2}$ \hspace{0.1cm}
Guang Li$^{2}$ \hspace{0.1cm}
Kun Ma$^{2}$ \hspace{0.1cm} \\
Long Chen$^{2}$ \hspace{0.1cm}
Zhongpu Xia$^{1}$ \hspace{0.1cm}
Qichao Zhang$^{1}$ \hspace{0.1cm}
Hangjun Ye$^{2}$ \hspace{0.1cm}
Dongbin Zhao$^{1}$
\vspace{0.2cm}
\\
{\small $^{1}$Institute of Automation, Chinese Academy of Sciences} {\small $^{2}$Xiaomi} \\
{\small $\dagger$Project Leader}
}
\begin{document}

\maketitle
\thispagestyle{empty}
\pagestyle{empty}

%%%%%%%%%%%%%%%%%%%%%%%%%%%%%%%%%%%%%%%%%%%%%%%%%%%%%%%%%%%%%%%%%%%%%%%%%%%%%%%%
\begin{abstract}
Current autonomous driving systems often favor end-to-end frameworks, which take sensor inputs like images and learn to map them into trajectory space via neural networks. Previous work has demonstrated that models can achieve better planning performance when provided with a prior distribution of possible trajectories. However, these approaches often overlook two critical aspects: 1) The appropriate trajectory prior can vary significantly across different driving scenarios. 2) Their trajectory evaluation mechanism lacks policy-driven refinement, remaining constrained by the limitations of one-stage supervised training. To address these issues, we explore improvements in two key areas. For problem 1, we employ MoE to apply different trajectory priors tailored to different scenarios. For problem 2, we utilize Reinforcement Learning to fine-tune the trajectory scoring mechanism. Additionally, we integrate models with different perception backbones to enhance perceptual features. Our integrated model achieved a score of 51.08 on the navsim ICCV benchmark, securing third place.
\end{abstract}


%%%%%%%%%%%%%%%%%%%%%%%%%%%%%%%%%%%%%%%%%%%%%%%%%%%%%%%%%%%%%%%%%%%%%%%%%%%%%%%%
\section{INTRODUCTION}

Frameworks like UniAD~\cite{hu2023uniad} utilize transformers with distinct functional modules to process and pass information, enabling mapping from sensor data to trajectory action space. Others, such as VAD~\cite{jiang2023vad}, adopt sparser perceptual representations to reduce computational demands. In recent end-to-end research, several works incorporate action anchors or priors. For instance, VADv2~\cite{chen2024vadv2}, and GTRS~\cite{li2025generalized} first cluster a large set of trajectories from the training dataset and then learn to select the optimal trajectory for the current scene from this cluster. DiffusionDrive~\cite{liao2025diffusiondrive} adds and removes noise based on a trajectory cluster, while GoalFlow~\cite{xing2025goalflow} clusters trajectory endpoints to serve as prior information for goal points. A key finding from recent methods is that incorporating a prior distribution over trajectories in some form helps the model generate safer and more reliable planned trajectories.

However, these methods typically rely on a fixed trajectory vocabulary or prior, despite the fact that action distributions can differ greatly across scenarios (e.g., going straight vs. turning). To address this limitation, we introduce Sparse MoE~\cite{shazeer2017outrageously} to differentiate between trajectory distributions. Specifically, trajectories from the vocabulary are routed to different experts based on the router's decision. Furthermore, we extend the standard supervised learning framework by performing additional fine-tuning using GRPO~\cite{shao2024deepseekmath}, following initial supervised learning on various driving score metrics. This aims to enhance the model's accuracy in predicting driving scores. Finally, building upon the GTRS baseline, we experiment with replacing different perception backbones and ensemble the outputs of our models. This approach led to a score of 51.08 on the navsimv2~\cite{navsimv2} ICCV competition, earning a third-place ranking.

In summary, our main contributions are as follows:

\begin{itemize}

\item We explore integrating Sparse MoE with a standard trajectory vocabulary, allowing trajectories to be processed by different experts to account for varying scenario-specific distributions.
\item We investigate augmenting supervised learning on driving scores with subsequent GRPO fine-tuning to improve the model's understanding and prediction of these scores.
\item We ensemble models with different perception backbones, achieving strong performance on the navsim ICCV benchmark.
\end{itemize}

\begin{figure*}[ht]
  \centering
  \includegraphics[width=1.0\textwidth]{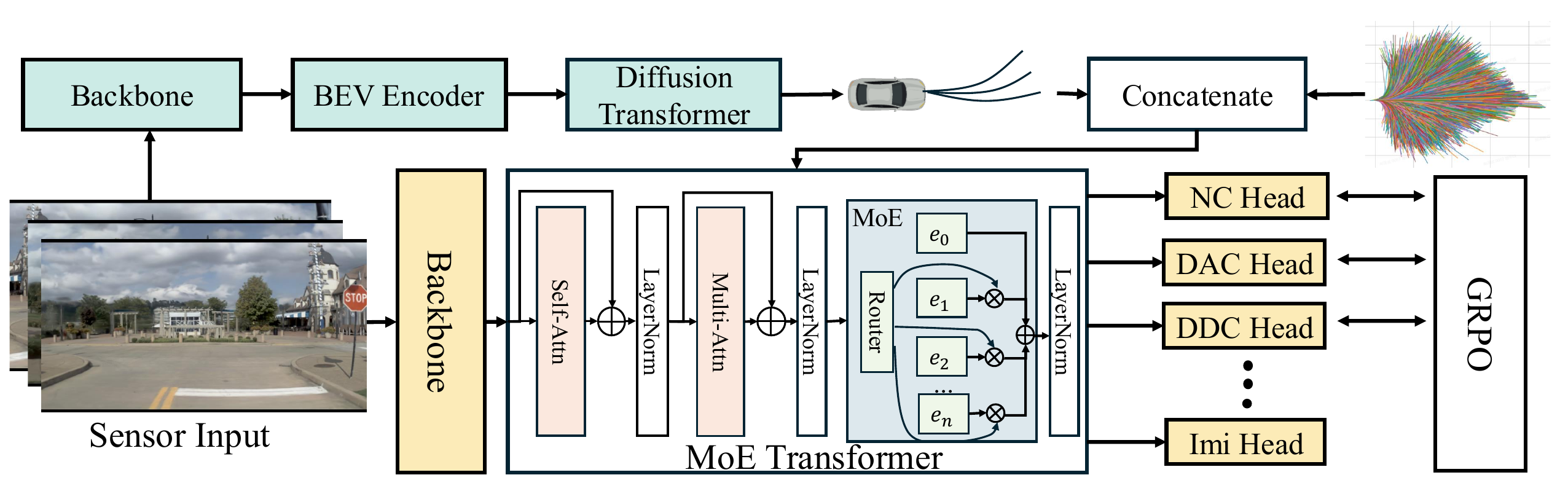}

  \caption{\textbf{Overview of the TrajMoE architecture.}}
  \label{fig:main_fig}
\end{figure*}

\section{METHODOLOGY}
\subsection{Base Model}
Based on the established anchor trajectory paradigm, our architectural foundation is built upon the state-of-the-art GTRS framework~\cite{li2025generalized}. The input to GTRS comprises a concatenated tensor $\mathrm{C} \in \mathbb{R}^{H \times W \times 3}$, formed by spatially stitching image data from the front, front-left, and front-right cameras. A powerful perception backbone, specifically V299~\cite{lee2019energy}, is employed to extract a dense and informative scene feature representation, denoted as $\mathrm{F}_{\text{cam}} \in \mathbb{R}^{T_{cam} \times D_{cam}}$, from this composite input.

Subsequently, a predefined set of trajectory anchors, termed the trajectory vocabulary $\mathbb{V}$, is embedded into a latent space $\mathrm{F}_{\mathrm{traj}}$. The camera feature $\mathrm{F}_{\text{cam}}$ and the trajectory vocabulary feature $\mathrm{F}_{\text{traj}}$ interact through a transformer architecture to produce an enhanced representation $\mathrm{F}_{\text{fusion}}$.

To facilitate discriminative scoring within the vocabulary, the GTRS model is optimized under a multi-task learning objective, supervising the predicted trajectory scores from several critical aspects.

\subsection{MoE Transformer}
In previous models, the distribution over the entire trajectory vocabulary was undifferentiated. We posit that different scenarios necessitate distinct trajectory priors, which means different vocabulary trajectory token needs to be processed via different network acrodding the scene input. To this end, we leverage MoE to achieve a more sufficient fusion between the embedding features of the trajectory vocabulary and the scene features $\mathrm{F_{cam}}$. Specifically, we employ a sparse MoE layer \cite{shazeer2017outrageously} to replace the standard FFN in the Transformer architecture. This sparse MoE layer consists of a router and N private experts and 1 shared expert. The router predicts a probability distribution over the N private experts for each input token and selects the top-k experts. The probabilities for these k experts are normalized via a softmax function and used as weights for a weighted sum of their outputs. The shared expert processes every input token and its output is directly added to the final result. Formally, for an input token representation x, the output y of the MoE layer is computed as follows:

\begin{equation}
    w(i;x) =  \frac{r_i(x)}{\sum_{j \in S} r_j(x)}, \quad i \in S
\end{equation}
\begin{equation}
    \mathrm{MoE}(x) = \sum_{i \in S} w(i;x) \cdot e_i(x) + e_0(x)
\end{equation}

where \( S \) is the set of selected top-k expert indices with \( |S| = k \), $x$ is the input token from the attention layer. \( r \) and \( e \) are router and experts separately. $k$ is the number of experts activated per token (where $k<N$). \( e_0 \) is the shared expert that processes every input token.

To prevent certain experts from being updated too frequently, which can lead to imbalanced specialization and potential performance degradation, we introduce an auxiliary load balancing loss during training. This loss incentivizes a more uniform distribution of routing assignments across experts, both in terms of the total gating weight (importance) and the number of tokens assigned (load). The Importance of each expert is $I_{e}$, which represents the sum of the gating weights assigned to each expert across all tokens in a batch. The Load of each expert is $L_{e}$, which represents the count of tokens assigned to each expert.
\begin{equation}
I_e = \sum_{i=1}^{N} P_{i,e}, \quad L_e = \sum_{i=1}^{N} \mathbbm{1}(P_{i,e} > 0)
\end{equation}
where $P$ is the $B\times E$ matrix of router probabilities after softmax. $\mathbbm{1}$ is the indicator function.

The balance loss for the layer is then defined as the sum of the squared coefficients of variation of the importance and load distributions:
\begin{equation}
\mathcal{L}_{\text{balance}} = \text{CV}^2(\mathbf{I}) + \text{CV}^2(\mathbf{L})
\end{equation}

Finally, the total balancing loss is obtained by summing the balancing losses from all MoE layers in the model:

\begin{table*}[!ht]
\centering
\footnotesize
\setlength{\tabcolsep}{3.5pt}
\begin{tabular}{l| c|c |c| c c c c c c c c c |c}
    \toprule
    Method 
    & \multicolumn{1}{c}{Img. Res.}
    & \multicolumn{1}{c}{Backbone}
    & Stage
    & NC
    & DAC
    & DDC
    & TLC
    & EP
    & TTC
    & LK 
    & HC 
    & EC 
    & EPDMS \\
    \midrule
    
    GTRS-Dense
    & \multirow{1}{*}{$512\times 2048$} 
    & \multirow{1}{*}{V2-99}
    & \shortstack{Stage 1 \\ Stage 2}
    & \shortstack{98.7 \\ \textbf{91.4}} 
    & \shortstack{95.8 \\ \textbf{89.2}} 
    & \shortstack{99.4 \\ \textbf{94.4}} 
    & \shortstack{99.3 \\ 98.8} 
    & \shortstack{72.8 \\ 69.5} 
    & \shortstack{98.7 \\ \textbf{90.1}} 
    & \shortstack{95.1 \\ 54.6} 
    & \shortstack{96.9 \\ 94.1} 
    & \shortstack{40.4 \\ 49.7} 
    & 41.7 \\
    \midrule

    \multirow{6}{*}{GTRS-Aug}
    & \multirow{1}{*}{$512\times 2048$} 
    & \multirow{1}{*}{V2-99}
    & \shortstack{Stage 1 \\ Stage 2}
    & \shortstack{\textbf{98.9} \\ 87.9} 
    & \shortstack{95.1 \\ 88.8} 
    & \shortstack{99.2 \\ 89.6} 
    & \shortstack{99.6 \\ 98.8} 
    & \shortstack{76.1 \\ 80.3} 
    & \shortstack{\textbf{99.1} \\ 86.0} 
    & \shortstack{94.7 \\ 53.5} 
    & \shortstack{\textbf{97.6} \\ 97.1} 
    & \shortstack{54.2 \\ 56.1} 
    & 42.1 \\
    \cmidrule{2-14}
    
    & \multirow{2}{*}{$256\times 1024$} 
    & \multirow{1}{*}{DINOv3-L}
    & \shortstack{Stage 1 \\ Stage 2}
    & \shortstack{98.3 \\ 88.0} 
    & \shortstack{\textbf{97.1} \\ 88.9} 
    & \shortstack{99.3 \\ 90.6} 
    & \shortstack{\textbf{99.7} \\ 98.1} 
    & \shortstack{77.4 \\ \textbf{84.9}} 
    & \shortstack{98.6 \\ 84.0} 
    & \shortstack{\textbf{96.0} \\ 55.4} 
    & \shortstack{97.5 \\ 96.9} 
    & \shortstack{54.2 \\ 52.5} 
    & 43.8 \\
    \cmidrule{3-14}
    
    & 
    & \multirow{1}{*}{DINOv3-L-M}
    & \shortstack{Stage 1 \\ Stage 2}
    & \shortstack{97.6 \\ 89.0} 
    & \shortstack{96.4 \\ 88.3} 
    & \shortstack{99.2 \\ 91.7} 
    & \shortstack{99.6 \\ 98.9} 
    & \shortstack{76.0 \\ 80.9} 
    & \shortstack{98.7 \\ 87.2} 
    & \shortstack{96.0 \\ \textbf{55.6}} 
    & \shortstack{97.3 \\ 96.8} 
    & \shortstack{\textbf{54.7} \\ 55.3} 
    & 43.1 \\
    \midrule
    
    TrajMoE-MoE
    & \multirow{1}{*}{$512\times 2048$} 
    & \multirow{1}{*}{V2-99}
    & \shortstack{Stage 1 \\ Stage 2}
    & \shortstack{98.8 \\ 90.0} 
    & \shortstack{96.6 \\ 88.0} 
    & \shortstack{\textbf{100.0} \\ 91.3} 
    & \shortstack{99.7 \\ 98.8} 
    & \shortstack{75.0 \\ 79.2} 
    & \shortstack{98.6 \\ 87.6} 
    & \shortstack{95.5 \\ 53.6} 
    & \shortstack{97.5 \\ 96.5} 
    & \shortstack{50.6 \\ \textbf{56.8}} 
    & 43.8\\  
    \midrule

    TrajMoE-GRPO
    & \multirow{1}{*}{$512\times 2048$} 
    & \multirow{1}{*}{V2-99}
    & \shortstack{Stage 1 \\ Stage 2}
    & \shortstack{98.0 \\ 88.0} 
    & \shortstack{96.6 \\ 87.9} 
    & \shortstack{99.2 \\ 90.8} 
    & \shortstack{99.3 \\ \textbf{99.0}} 
    & \shortstack{\textbf{77.3} \\ 82.5} 
    & \shortstack{98.0 \\ 85.3} 
    & \shortstack{96.0 \\ 52.4} 
    & \shortstack{97.5 \\ \textbf{97.1}} 
    & \shortstack{54.2 \\ 51.2} 
    & 42.7\\  
    
    \midrule
    \midrule

    GTRS-Aug
    & \multirow{1}{*}{$512\times 2048$} 
    & \multirow{1}{*}{V2-99+VGGT}
    & \shortstack{Stage 1 \\ Stage 2}
    & \shortstack{95.7 \\ 86.5} 
    & \shortstack{\textbf{98.6} \\ 93.8} 
    & \shortstack{98.6 \\ 94.5} 
    & \shortstack{\textbf{98.6} \\ \textbf{96.9}} 
    & \shortstack{\textbf{85.6} \\ \textbf{85.3}} 
    & \shortstack{97.9 \\ 83.2} 
    & \shortstack{\textbf{93.6} \\ \textbf{57.0}} 
    & \shortstack{94.3 \\ 98.3} 
    & \shortstack{57.1 \\ 62.8} 
    & 47.0 \\
    \midrule

    GTRS+TrajMoE
    & * 
    & *
    & \shortstack{Stage 1 \\ Stage 2}
    & \shortstack{\textbf{96.4} \\ \textbf{87.5}} 
    & \shortstack{99.2 \\ \textbf{96.5}} 
    & \shortstack{\textbf{100.0} \\ \textbf{97.0}} 
    & \shortstack{98.5 \\ 96.6} 
    & \shortstack{85.6 \\ 84.2} 
    & \shortstack{\textbf{99.2} \\ \textbf{86.3}} 
    & \shortstack{93.5 \\ 55.4} 
    & \shortstack{\textbf{95.0} \\ \textbf{98.9}} 
    & \shortstack{\textbf{70.0} \\ \textbf{74.7}} 
    & \textbf{51.0}\\
    \bottomrule
\end{tabular}
\caption{\textbf{Performance on the navhard and private-hard of NAVSIMv2~\cite{navsimv2}.}}
\label{table:result}
\end{table*}

\subsection{RL-finetune}
We adopt the Group Relative Policy Optimization (GRPO) strategy to fine-tune the scoring heads of GTRS. 
For each scoring head, the model outputs a predicted score $\hat{s}$, parameterized as a Gaussian distribution, from which a group of $N$ candidate scores is sampled to encourage exploration:
\begin{equation}
    \hat{s} \sim \mathcal{N}(\mu_\theta, \sigma_\theta^2), \quad 
    \{s_i\}_{i=1}^N \sim \mathcal{N}(\mu_\theta, \sigma_\theta^2),
\end{equation}
where $\mu_\theta$ and $\sigma_\theta$ denote the mean and variance estimated by the policy network $\pi_\theta$.  

Each sampled score $s_i$ is assigned a reward based on its deviation from the ground-truth score $s^{\ast}$, which is then normalized within the group to remove scale sensitivity:
\begin{equation}
    r_i = -|s_i - s^{\ast}|, \quad
    \tilde{r}_i = \frac{r_i - \text{mean}(\mathbf{r})}{\text{std}(\mathbf{r})}, \;\; 
    \mathbf{r} = \{r_1, r_2, \dots, r_N\}.
\end{equation}

The normalized rewards are accumulated to compute the relative advantage of each candidate:
\begin{equation}
    \text{Adv}_i = \sum_{r \ge i} \tilde{r}_i,
\end{equation}

The policy objective is then defined as a clipped surrogate function with KL regularization:
\begin{equation}
\begin{aligned}
    J(\theta) = \frac{1}{N}\sum_i \Big\{ 
        &\min \big[f_1, f_2 \big] 
        - \beta D_{\text{KL}}(\pi_\theta \| \pi_{\theta_{\text{ref}}}) 
    \Big\}, \\
    f_1 &= \left(\frac{\pi_\theta(s_i)}{\pi_{\theta_{\text{old}}}(s_i)}\right) \text{Adv}_i, \\
    f_2 &= \text{clip}\left(\frac{\pi_\theta(s_i)}{\pi_{\theta_{\text{old}}}(s_i)}, 1 - \varepsilon, 1 + \varepsilon\right) \text{Adv}_i,
\end{aligned}
\end{equation}
where $\pi_{\theta_{\text{old}}}$ is the policy from the previous iteration, and $\pi_{\theta_{\text{ref}}}$ is the reference policy. The KL divergence is defined as
\begin{equation}
    D_{\text{KL}} = \tfrac{1}{2} \Big[ \log \sigma_\theta^2 - \log \sigma_{\text{ref}}^2 
    + \tfrac{\sigma_{\text{ref}}^2 + (\mu_\theta - \mu_{\text{ref}})^2}{\sigma_\theta^2} - 1 \Big],
\end{equation}
where $(\mu_{\text{ref}}, \sigma_{\text{ref}})$ are the parameters of $\pi_{\theta_{\text{ref}}}$.  

Finally, the RL fine-tuning loss is formulated as:
\begin{equation}
    \mathcal{L}_{\text{RL}} = -J(\theta) + \lambda D_{\text{KL}},
\end{equation}
where $\lambda$ balances policy improvement with KL regularization.  

Through this reinforcement learning fine-tuning procedure, each scoring head adaptively learns an optimal scoring distribution conditioned on perception inputs, thereby improving alignment with ground-truth supervision while ensuring robustness under uncertainty and exploration noise.

\subsection{Model Ensembling}
Based on the GTRS-Aug framework, we conduct a series of extension experiments~\ref{table:result} to enhance the model's perceptual capabilities and planning performance. To improve scene understanding, we explore various perception backbones:
For the V2-99+VGGT configuration, we retain the original V2-99 perception network while integrating depth-aware features through feature-level addition with VGGT~\cite{wang2025vggt} outputs, thereby enriching the perceptual representation with geometric information.
We also investigate DINOv3-L~\cite{simeoni2025dinov3} based architectures, where both DINOv3-L and DINOv3-L-M variants employ the DINOv3-L framework, with the latter specifically incorporating motion blur data augmentation during training to enhance robustness.
Beyond backbone modifications, we introduce two algorithmic enhancements: TrajMoE-MoE integrates a Mixture of Experts mechanism into GTRS-Aug to enable scenario-adaptive trajectory processing, while TrajMoE-GRPO further incorporates GRPO-based fine-tuning to refine trajectory scoring through reinforcement learning.

Finally, we ensemble trajectories from all these models on GTRS+TrajMoE via weighted averaging, leveraging complementary strengths across different architectural designs and training strategies to achieve superior overall planning performance.

\section{EXPERIMENT}
\subsection{Main Result.}
Our experimental results are summarized in Table~\ref{table:result}. It is worth noting that all compared methods employ the same evaluation protocol with identical metric compositions. Specifically, GTRS+TrajMoE and GTRS-Aug which based on the V2-99 and VGGT is evaluated on the private-test-hard split, which serves as the official competition benchmark and is conducted on the evaluation server. In contrast, other methods are tested on the navhard split, a locally administered test set that shares the same metric configuration as the official benchmark.


\begin{thebibliography}{99}
\bibitem{navsimv2} W.~Cao, M.~Hallgarten, T.~Li, D.~Dauner, X.~Gu, C.~Wang, Y.~Miron, M.~Aiello, H.~Li, I.~Gilitschenski \emph{et al.}, ``Pseudo-simulation for autonomous driving,'' \emph{arXiv preprint arXiv:2506.04218}, 2025.

\bibitem{hu2023uniad} Y.~Hu, J.~Yang, L.~Chen, K.~Li, C.~Sima, X.~Zhu, S.~Chai, S.~Du, T.~Lin, W.~Wang \emph{et al.}, ``Planning-oriented autonomous driving,'' in \emph{CVPR}, 2023.

\bibitem{jiang2023vad} B.~Jiang, S.~Chen, Q.~Xu, B.~Liao, J.~Chen, H.~Zhou, Q.~Zhang, W.~Liu, C.~Huang, and X.~Wang, ``Vad: Vectorized scene representation for efficient autonomous driving,'' in \emph{ICCV}, 2023.

\bibitem{chen2024vadv2} S.~Chen, B.~Jiang, H.~Gao, B.~Liao, Q.~Xu, Q.~Zhang, C.~Huang, W.~Liu, X.~Wang \emph{et al.}, ``Vadv2: End-to-end vectorized autonomous driving via probabilistic planning,'' \emph{arXiv preprint arXiv:2402.13243}, 2024.

\bibitem{liao2025diffusiondrive} B.~Liao, S.~Chen, H.~Yin, B.~Jiang, C.~Wang, S.~Yan, X.~Zhang, X.~Li, Y.~Zhang, Q.~Zhang \emph{et al.}, ``Diffusiondrive: Truncated diffusion model for end-to-end autonomous driving,'' in \emph{CVPR}, 2025.

\bibitem{xing2025goalflow} Z.~Xing, X.~Zhang, Y.~Hu, B.~Jiang, T.~He, Q.~Zhang, X.~Long, and W.~Yin, ``Goalflow: Goal-driven flow matching for multimodal trajectories generation in end-to-end autonomous driving,'' in \emph{CVPR}, 2025.

\bibitem{lee2019energy} Y.~Lee, J.~Hwang, S.~Lee, Y.~Bae, J.~Park \emph{et al.}, ``An Energy and GPU-Computation Efficient Backbone Network for Real-Time Object Detection,'' \emph{CVPR}, 2019.

\bibitem{shao2024deepseekmath} Z.~Shao, P.~Wang, Q.~Zhu, R.~Xu, J.~Song, X.~Bi, H.~Zhang, M.~Zhang, Y.~Li, Y.~Wu \emph{et al.}, ``Deepseekmath: Pushing the limits of mathematical reasoning in open language models,'' \emph{arXiv preprint arXiv:2402.03300}, 2024.

\bibitem{li2025generalized} Z.~Li, W.~Yao, Z.~Wang, X.~Sun, J.~Chen, N.~Chang, M.~Shen, Z.~Wu, S.~Lan, J.~Alvarez \emph{et al.}, ``Generalized Trajectory Scoring for End-to-end Multimodal Planning,'' \emph{arXiv preprint arXiv:2506.06664}, 2025.

\bibitem{shazeer2017outrageously} N.~Shazeer, A.~Mirhoseini, K.~Maziarz, A.~Davis, Q.~Le, G.~Hinton, J.~Dean \emph{et al.}, ``Outrageously large neural networks: The sparsely-gated mixture-of-experts layer,'' \emph{arXiv preprint arXiv:1701.06538}, 2017.

\bibitem{simeoni2025dinov3} O.~Sim\'eoni, H.~Vo, M.~Seitzer, F.~Baldassarre, M.~Oquab, C.~Jose, V.~Khalidov, M.~Szafraniec, S.~Yi, M.~Ramamonjisoa \emph{et al.}, ``DINOv3,'' \emph{arXiv preprint arXiv:2508.10104}, 2025.

\bibitem{wang2025vggt} J.~Wang, M.~Chen, N.~Karaev, A.~Vedaldi, C.~Rupprecht, D.~Novotny \emph{et al.}, ``Vggt: Visual geometry grounded transformer,'' \emph{CVPR}, pp. 5294--5306, 2025.
\end{thebibliography}
\end{document}